# Current Landscape of the Russian Sentiment Corpora

**Kotelnikov E. V.**
Vyatka State University, Kirov, Russia;
ITMO University, Saint Petersburg, Russia
kotelnikov.ev@gmail.com

**Abstract**

Currently, there are more than a dozen Russian-language corpora for sentiment analysis, differing in the source of the texts, domain, size, number and ratio of sentiment classes, and annotation method. This work examines publicly available Russian-language corpora, presents their qualitative and quantitative characteristics, which make it possible to get an idea of the current landscape of the corpora for sentiment analysis. The ranking of corpora by annotation quality is proposed, which can be useful when choosing corpora for training and testing. The influence of the training dataset on the performance of sentiment analysis is investigated based on the use of the deep neural network model BERT. The experiments with review corpora allow us to conclude that on average the quality of models increases with an increase in the number of training corpora. For the first time, quality scores were obtained for the corpus of reviews of ROMIP seminars based on the BERT model. Also, the study proposes the task of the building a universal model for sentiment analysis.

# Текущее состояние русскоязычных корпусов для анализа тональности текстов

**Котельников Е. В.**
Вятский государственный университет, Киров, Россия;
Университет ИТМО, Санкт-Петербург, Россия
kotelnikov.ev@gmail.com

**Аннотация**

В настоящее время существует более десятка русскоязычных корпусов для анализа тональности, отличающихся источником текстов, предметной областью, размерами, количеством и соотношением классов тональности, способом разметки. В работе рассматриваются общедоступные русскоязычные корпуса, приводятся их качественные и количественные характеристики, позволяющие составить представление о текущем состоянии корпусов для анализа тональности. Предлагается ранжирование корпусов по качеству разметки, которое может быть полезно при выборе корпусов для обучения и тестирования. Исследуется влияние обучающей выборки на качество анализа тональности на основе применения глубокой нейросетевой модели BERT. Эксперименты с корпусами отзывов позволяют сделать вывод о том, что при увеличении количества обучающих корпусов качество моделей в среднем повышается. Впервые получены оценки качества для корпусов отзывов семинаров РОМИП на основе модели BERT. Также ставится задача построения универсальной модели для анализа тональности.

## 1 Introduction

Currently, the text sentiment analysis is still an urgent problem. Despite the fact that modern deep neural network models allow for some datasets to reach an accuracy close to 100% in the case of binary classification (positive/negative) of the SST-2 English movie review corpus [8], with the number of classes more than two the accuracy does not exceed 60% [21].

The most important factor in the construction of sentiment analysis systems is the availability of a variety of high-quality text corpora. The English corpora for sentiment analysis have been fairly well researched [25]. The first Russian-language text corpora devoted to sentiment analysis appeared in 2011. These are three corpora of reviews for books, movies and cameras prepared for the ROMIP (Russian Information Retrieval Evaluation Seminar) competition [4]. Over the past 10 years, more than a dozen Russian-language corpora have been annotated by sentiment and made available for public access.

Russian-language corpora, as opposed to English-language, despite some recent works [7, 20], have not been sufficiently researched yet. In particular, there are no works devoted to the analysis of the quality of corpora, as well as studies of the dependence of the models' quality on the training corpora. There are also no performance scores of the modern deep neural network models for the review corpora of the ROMIP competitions.

The most important characteristics of the corpora intended for sentiment analysis are the source of texts, the domain, the size of the corpus, the size of the texts, the number and ratio of sentiment classes, the annotation method, the presence of a split into training and test parts. This paper examines the existing Russian-language publicly available corpora, annotated by sentiment.

The contribution of this work is as follows:

- an overview of all publicly available Russian-language corpora with detailed characteristics is provided;
- the ranking of corpora by annotation quality is proposed;
- new quality ratings have been obtained for the existing Russian-language corpora of reviews;
- the influence of the training dataset on the performance of the sentiment analysis of reviews was investigated.

The study also proposes the task of the building a universal model for sentiment analysis.

The rest of the paper is organized as follows. The second section reveals the characteristics of existing corpora. The third section is devoted to the materials and methods used in the experimental study. In the fourth section, the results of the experiments are presented and discussed. The fifth section provides an overview of previous works on Russian-language corpora for sentiment analysis. In the final section the conclusions are drawn and directions for further research are indicated.

## 2 The Russian text corpora for sentiment analysis

### 2.1 Characteristics of corpora

This section discusses existing Russian-language corpora for sentiment analysis. As noted in the Introduction, the main characteristics of the corpora are the source of the texts, the domain, the size of the corpus, the size of the texts, the number and ratio of sentiment classes, the annotation method, the presence of a split into training and test parts.

*Sources of the texts*. All corpora can be divided into four sources of the texts: 1) reviews of products, works of art and organizations; 2) tweets; 3) posts on social networks; 4) news articles. The source of the texts determines the domain, style and size of the texts.

*The domain* defines the topic of the texts, for example, restaurant reviews or political news. There are corpora without an explicitly defined domain (for example, RuTweetCorp).

*Corpus sizes* vary from several dozen (RuSentRel) to hundreds of thousands of texts (RuTweetCorp). *The size of the texts* is related to the source and ranges from a few words (for tweets) to several thousand words (for reviews, news and social media posts).

*Sentiment classes.* A one-dimensional scale is often used to indicate sentiment (Figure 1a). There are 6 main classes: positive, weakly positive, negative, weakly negative, neutral and contradictory. However, in the case of the one-dimensional scale, uncertainty arises with an intermediate (zero) value, which can have the meaning of a neutral sentiment (the absence of sentiment) or contradictory (presence of both positive and negative sentiments). Therefore, it is more convenient to represent the sentiment classes on a plane (the boundaries between the classes are shown conditionally; the symmetry of the positive and negative sentiment is assumed) (Figure 1b).

The minimum number of classes in existing corpora is two – positive and negative (hereinafter referred to as "+" and "–"). In the case of three-class annotation, the third class is considered either contradictory (referred to as “±”) or neutral (referred to as “0”), and these two classes are not always separated in the corpora. There are also cases with a five-point rating (from 1 to 5), where the value "3" can mean either contradictory or neutral class.

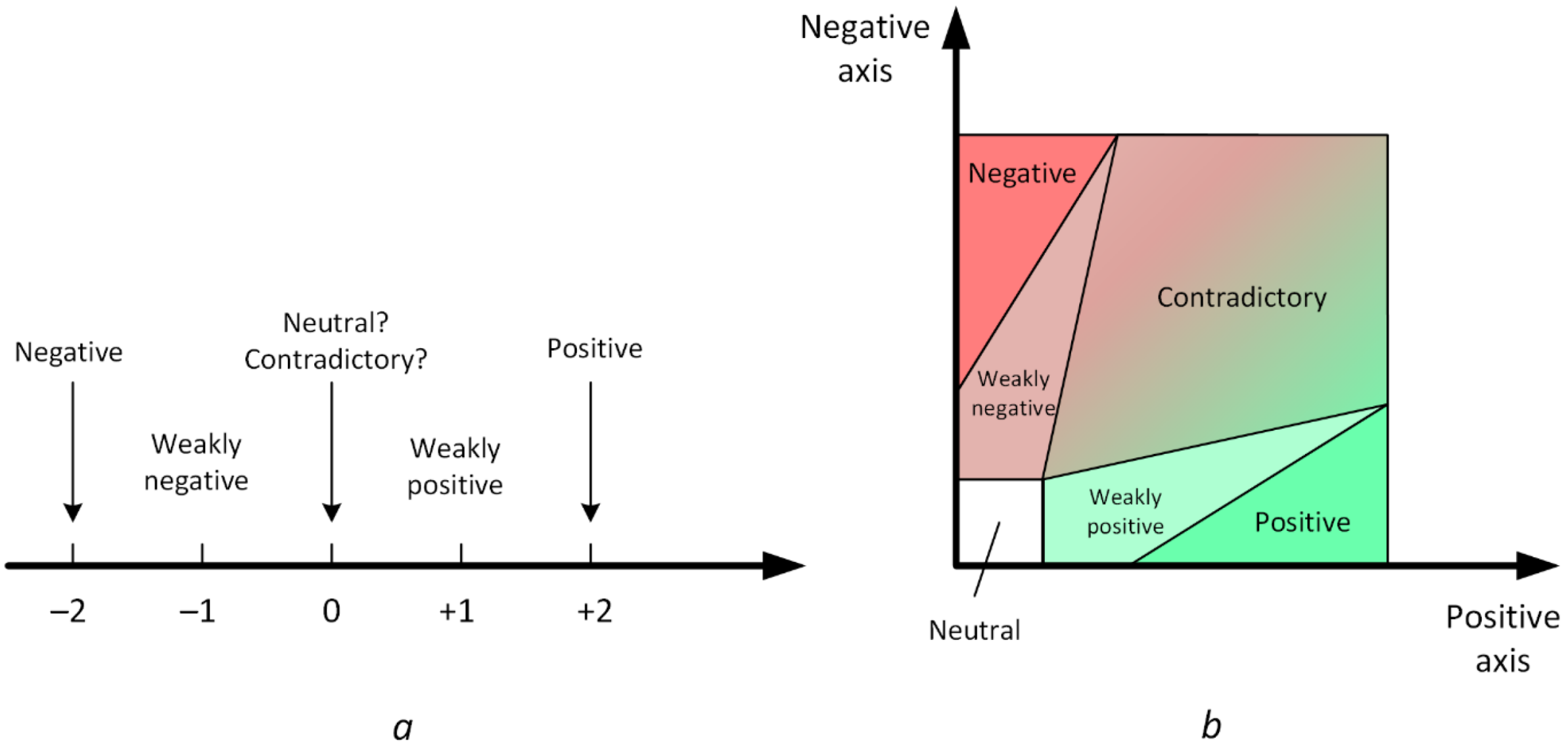

Figure 1: Sentiment spaces: a – 1D space, b – 2D space

*The ratio of the sentiment classes.* In many domains, there is a natural imbalance of texts according to the sentiment classes, which turns (if no special measures were taken) into corpora. For example, reviews about products, works of art and organizations are characterized by a significant predominance of positive texts, while tweets that mention banks and telecommunications companies tend to be more negative. It is known that class imbalance degrades the performance of classification [2], but there are methods that can level this aspect [23].

*Annotation method.* Existing corpora are annotated using three main approaches – manual annotation, automatic annotation, and use of the author's annotation. Manual annotation, in turn, is divided into expert annotation, when texts are marked up by a small number of qualified and motivated annotators, and crowdsourced annotation, in which a large number of crowdworkers are involved for annotation on a paid or free basis using specialized web platforms. Automatic annotation uses indirect sentiment information available in the text, for example, emoticons. In the third approach, the sentiment class of texts (usually reviews) is indicated in accordance with the score provided by the author of the text.

An important issue is the degree of confidence in the quality of the annotation. We propose to distinguish the following confidence levels, depending on the number of annotators and the approach to annotation:

1. High level (L1) – the annotation was carried out by at least two annotators, including on the basis of a crowdsourcing approach;
2. Middle level (L2) – the annotation was carried out by only one annotator;
3. Lower middle level (L3) – the annotation was based on the author's score;
4. Low level (L4) – the annotation was carried out automatically.

*Train/test split.* In many existing corpora there is a split into training and test parts – this is important for the reproducibility of experimental results.

### 2.2 Existing corpora

Table 1 shows the qualitative characteristics of existing corpora for sentiment analysis; Table 2 presents the quantitative characteristics of these corpora.

**ROMIP-2011**. For the competition of sentiment analysis systems within the ROMIP-2011 seminar, three corpora were created – reviews of books, movies and cameras [4]. Reviews were collected using queries on Yandex's Blog Search. Each corpus included two parts: a training part, which was marked up using the author's score, and a test part, which was marked up by two annotators. The author's scores are given on a scale of [1..10] for reviews about books and movies and [1..5] for reviews about cameras. Test annotation was performed for three scales: binary (positive/negative), three-class (adding a contradictory sentiment) and five-class [1..5]. In the original paper [4], the results of systems for test reviews were evaluated according to the AND scheme (the system's score coincides with the score of both annotators) and OR scheme (the system score coincides with the score of at least one annotator). In this paper the AND scheme is used.

| **Corpus** | **Source** | **Domain** | **Annotation** | **Confidence level** | **Number of classes: labels** |
|---|---|---|---|---|---|
| ROMIP-2011 | Reviews | Book, movie and camera reviews | Train: author<br>Test:<br>2 annotators | Train: L3<br>Test: L1 | 2: {+, –}<br>3: {+, –, ±}<br>5: {1, 2, 3, 4, 5} |
| ROMIP-2012 (reviews) | Reviews | Book, movie and camera reviews | 1 annotator | Test: L2 | 2: {+, –}<br>3: {+, –, ±}<br>5: {1, 2, 3, 4, 5} |
| ROMIP-2012 (quotes) | News | – | 1 annotator | Train: L2<br>Test: L2 | Train, 4: {+, –, ±, 0}<br>Test, 3: {+, –, 0} |
| SentiRuEval-2015 (reviews) | Reviews | Car and restaurant reviews | 1 annotator (+checking) | Train: L2<br>Test: L2 | 4: {+, –, ±, 0} |
| SentiRuEval-2015 (tweets) | Twitter | Banks, telecom companies | 3 annotators | Train: L1<br>Test: L1 | Train, 4: {+, –, ±, 0}<br>Test, 3: {+, –, 0} |
| SentiRuEval-2016 | Twitter | Banks, telecom companies | Crowdsourcing | Test: L1 | 3: {+, –, 0} |
| SemEval-2016 | Reviews | Restaurant reviews | 2 annotators | L1 | 4: {+, –, ±, 0} |
| LinisCrowd | Social media posts | – | Crowdsourcing:<br>1 annotator,<br>>1 annotator | 1 annotator: L2<br>>1 annotator: L1 | 5: {–2, –1, 0, 1, 2} |
| Russian Hotel Reviews | Reviews | Hotel reviews | Author | Train: L3<br>Test: L3 | 5: {1, 2, 3, 4, 5} |
| RuSentiment | Social media posts | – | 3 annotators | Train: L1<br>Test: L1 | 3: {+, –, 0} |
| RuSentRel | News | International politics | 2 annotators | Train: L1<br>Test: L1 | 2: {+, –} |
| RuReviews | Reviews | Woman clothes and accessories reviews | Author | L3 | 3: {+, –, 0} |
| RuTweetCorp | Twitter | – | Automatic | L4 | 2: {+, –} |
| Kaggle Russian News Dataset | News | Kazakh news | ? | ? | 3: {+, –, 0} |
| Twitter Sentiment for 15 European Languages | Twitter | – | 1 annotator | L2 | 3: {+, –, 0} |

Table 1: Qualitative characteristics of text corpora

| Corpus | Domain | Number of texts | Training set | Test set | pos/neg/third class, % | Mean number of words (± Std Dev) |
|---|---|---|---|---|---|---|
| ROMIP-2011 (2 classes) | Books | 19,946 | 19,680 | 266 | 89.6/10.4/0.0 | 49.3±101.6 |
| | Movies | 12,653 | 12,341 | 312 | 84.7/15.3/0.0 | 77.9±161.4 |
| | Cameras | 8,873 | 8,618 | 255 | 88.2/11.8/0.0 | 52.0±76.8 |
| ROMIP-2012 (reviews, 2 classes) | Books | 129 | – | 129 | 86.8/13.2/0.0 | 199.9±319.9 |
| | Movies | 408 | – | 408 | 80.9/19.1/0.0 | 298.5±545.0 |
| | Cameras | 411 | – | 411 | 96.6/3.4/0.0 | 57.0±70.6 |
| ROMIP-2012 (quotes, 2 classes) | – | 8,833 | 4,260 | 4,573 | 29.0/42.5/28.5 | 35.3±24.8 |
| SentiRuEval-2015 (reviews) | Cars | 403 | 203 | 200 | 52.9/13.9/33.3 | 116.6±68.2 |
| | Restaurants | 403 | 200 | 203 | 70.0/13.4/16.6 | 132.6±44.2 |
| SentiRuEval-2015 (tweets) | Banks | 9,417 | 4,883 | 4,534 | 7.4/18.2/74.4 | 9.6±4.9 |
| | Telecom | 8,613 | 4,839 | 3,774 | 14.5/28.2/57.3 | 12.2±5.5 |
| SentiRuEval-2016 | Banks | 3,302 | – | 3,302 | 9.1/23.1/67.8 | 12.3±4.9 |
| | Telecom | 2,198 | – | 2,198 | 8.3/45.9/45.9 | 14.4±5.4 |
| SemEval-2016 | Restaurants | 405 | 302 | 103 | 72.1/13.1/14.8 | 133.5±44.7 |
| LinisCrowd (1 annotator) | – | 28,853 | – | – | 7.7/42.5/49.8 | 148.6±103.6 |
| LinisCrowd (>1 annotator) | – | 10,566 | – | – | 6.9/40.4/52.7 | 139.8±75.9 |
| Russian Hotel Reviews | Hotels | 57,204 | 50,328 | 6,876 | 82.8/6.1/11.1 | 92.6±103.4 |
| RuSentiment | – | 26,745 | 24,124 | 2,621 | 37.8/14.6/47.6 | 12.6±16.9 |
| RuReviews | Woman clothes and accessories | 89,999 | – | – | 33.3/33.3/33.3 | 20.2±19.9 |
| RuTweetCorp | – | 226,834 | – | – | 50.7/49.3/0.0 | 12.2±4.9 |
| Kaggle Russian News Dataset | Kazakhstan news | 8,263 | 8,263 | – | 33.8/17.4/48.8 | 520.2±1192.2 |

Table 2: Quantitative characteristics of text corpora
(“third class” – neutral and/or contradictory class; empty texts are excluded)

**ROMIP-2012**. The ROMIP-2012 competition was held in 2012 [5]. The corpora of ROMIP-2011 were used as training datasets. To obtain test data, new corpora of reviews about books, movies and cameras were marked up with a single annotator. The same scales were used for annotation as in ROMIP-2011 – 2-, 3- and 5-class. In addition to reviews, training and test corpora of quotes from news were also prepared for the competition. The scale {+, –, ±, 0} was used for annotation of the training corpus; there was no contradictory sentiment in the annotation of the test corpus.

**SentiRuEval-2015**. In 2015 the next sentiment analysis systems competition took place, which was aimed at two tasks: aspect-based sentiment analysis of reviews and object-oriented sentiment analysis of tweets [11]. For the first task, training and test corpora of car reviews were prepared, annotated by the aspects of Drivability, Reliability, Safety, Appearance, Comfort, Costs and General, and reviews about restaurants, for which the aspects of Food, Service, Interior, Price and General were highlighted. The annotation on the scale {+, –, ±, 0} was carried out by one annotator, but then a check was carried out. Table 2 provides data on the General aspect.

For the second task, training and test corpora of tweets about eight banks and seven telecommunications companies were annotated. The markup was done by three annotators; a voting scheme was used to obtain the final score. The scales {+, –, ±, 0} and {+, –, 0} were used for the training and test data respectively.

**SentiRuEval-2016**. The SentiRuEval-2016 competition also analyzed tweets in relation to banks and telecommunications companies [12]. The training corpora were built by combining the training and test corpora of the SentiRuEval-2015. Crowdsourcing was used to annotate the test data on the {+, –, 0} scale.

**SemEval-2016**. In 2016 within the international competition SemEval-2016, the subtask of aspect-based sentiment analysis, including Russian-language reviews of restaurants, was singled out [15]. The training corpus was built on the basis of the corresponding SentiRuEval-2015 corpus and more than half overlaps with it. The test corpus was built from scratch. The annotation on the scale {+, –, ±, 0} was carried out by two annotators.

**LinisCrowd**. Within the Linis Crowd project, posts and comments of Top-2000 bloggers on LiveJournal were offered for crowdsourcing annotation [9]. The scale was {–2, –1, 0, 1, 2}. For each text a different number of scores were obtained – from 1 to 57. We divided the corpus into two parts: texts annotated by only one user (middle confidence level – L2) and texts annotated by several users (high confidence level – L1). In texts with several scores the final sentiment score was chosen according to the majority of scores.

**Russian Hotel Reviews**. Rybakov and Malafeev [18] offered a corpus of hotel reviews collected from tripadvisor.ru. The scores of reviews' authors were used. These scores correspond to a five-point scale for the aspects of Price-quality ratio, Location, Room, Cleanliness, Service, Quality of sleep and General. The corpus is divided into training and test parts.

**RuSentiment**. The corpus of posts on the social network VKontakte was presented in [16]. The corpus was marked by three annotators on the {+, –, 0} scale (the positive subcategory Speech Act was also highlighted, which we included in the positive class).

**RuSentRel**. Loukachevitch and Rusnachenko [13] presented a corpus of news articles on international politics from the inosmi.ru website. This corpus is annotated in relation to the named entities mentioned in the texts. The annotation was carried out on a {+, –} scale by two annotators; the third annotator resolved the contradictions. An overall sentiment score of the text was not made, therefore, information on the corpus is not provided in Table 2.

**RuReviews**. Smetanin and Komarov [19] presented a corpus of reviews about women's clothes and accessories, collected from some major e-commerce site. The original five-point scale was transformed by the authors into a three-point scale {+, –, 0}.

**RuTweetCorp**. Rubtsova [17] presented the largest corpus of Russian-language tweets for sentiment analysis to date. The annotation on the {+, –} scale was carried out automatically based on emoticons in the tweets.

**Kaggle Russian News Dataset**. The Kaggle[1] website presents a corpus of Kazakhstan news in Russian, annotated on a {+, –, 0} scale. The source of the texts and the method of annotation are unknown. The training and test parts of the corpus are available on the Kaggle website, but the sentiment scores are given only for the training part.

**Twitter Sentiment for 15 European Languages**. Mozetic et al. [14] consider tweet corpora for 15 European languages, including Russian. 93,321 messages were annotated with one annotator on the {+, –, 0} scale. However, the tweets themselves are not available (only their IDs are available), so Table 2 does not provide information about the corpus.

# 3 Materials and methods

## 3.1 BERT

To classify texts by sentiment, we use a deep neural network language model BERT (Bidirectional Encoder Representations from Transformers) [6], which showed the best results for the sentiment analysis in Russian [7, 20].

BERT is a Transformer encoder [22], which includes multiple layers; each layer contains a self-attention mechanism. Devlin et al. [6] presented two versions of BERT – $\text{BERT}_{\text{BASE}}$ and $\text{BERT}_{\text{LARGE}}$. In the base version the number of layers is 12, in the large version – 24. Work with BERT, as a rule, involves two stages. At the first stage, a language model is built by training on the tasks of predicting masked words and the next sentence using large text corpora (for example, Wikipedia). At the second stage, the pre-trained language model is fine-tuned to a specific task, for example, sentiment analysis. The impressive BERT results are based on the deep bi-directionality of the model, that is, considering left and right word contexts across all layers.

[1] https://www.kaggle.com/c/sentiment-analysis-in-russian.

BERT uses subword tokenization to represent input texts [24]. The maximum input size for BERT is 512 tokens (subwords). In addition to word tokens, special tokens are used, for example, [CLS], which is always placed first and represents the text as a whole.

To classify texts in BERT a linear layer with a SoftMax function is used. The weights of this layer are randomly initialized before fine-tuning. This layer receives as the input the output vector corresponding to the special token [CLS].

In our work as a pre-trained language model we have used the Russian-language version of BERT – RuBERT, proposed by Kuratov and Arkhipov [10]. To train this model, a multilingual version of the $\mathrm{BERT_{BASE}}$ was taken (12 layers, hidden size 768, feed-forward hidden size 3,072, and 12 self-attention heads). This version was retrained on the Russian-language Wikipedia and news corpus.

### 3.2 Corpora

One of the main goals of our work is to study the dependence of the performance of sentiment analysis on training data. To do this, we took 7 corpora of reviews as training datasets: three train parts of the ROMIP-2011 corpora (referred to as *R11_book_tr*, *R11_mov_tr* and *R11_cam_tr*), two train parts of the SentiRuEval-2015 corpora (*SRV15_car_tr* and *SRV15_rest_tr*), RuReviews corpus (*RuReviews*) and train part of the Russian Hotel Reviews (*Hotel_tr*). As test datasets we selected 9 review corpora: three test parts of the ROMIP-2011 corpora (*R11_book_te*, *R11_mov_te* and *R11_cam_te*), three test parts of the ROMIP-2012 corpora (*R12_book_te*, *R12_mov_te* and *R12_cam_te*), two test parts of the SentiRuEval-2015 corpora (*SRV15_car_te* and *SRV15_rest_te*) and test part of the Russian Hotel Reviews (*Hotel_te*).

Training corpora have the confidence levels L2 (2 corpora of the SentiRuEval-2015) and L3 (the remaining 5 corpora); test corpora – L1 (ROMIP-2011), L2 (ROMIP-2012, SentiRuEval-2015) and L3 (Russian Hotel Reviews).

Versions with binary scores (positive/negative) have been taken for all the corpora, that is, the task of two-class classification was solved. The characteristics for all of these corpora are shown in Tables 1 and 2.

The following preprocessing procedures were applied to the texts:

- URLs, e-mails and phone numbers were replaced with special tokens;
- characters that were repeated more than two times were replaced with a sequence of two such characters;
- joyful and sad emoticons were replaced with “joy” and “sadness” tokens.

The input size of the BERT model is limited to 512 tokens; the length of reviews often exceeds this size (see Table 2). To work with long reviews, the following strategy was used: half of the input tokens were taken from the beginning of the text, the other half – from the end of the text. This strategy is based on the fact that the main opinion in a review is often given either at the beginning or at the end.

## 4 Results and discussion

### 4.1 Experimental design

To select hyperparameters a preliminary series of experiments was carried out using only training corpora. As a result, the following hyperparameters values were selected, which were used in all subsequent experiments: the number of epochs is 5, the batch size is 8, and the learning rate is 2e–5.

The experiments were carried out using the Google Colab Pro platform, which provides graphics cards Tesla V100-SXM2-16GB or Tesla P100-PCIE-16GB.

Since the BERT training process is stochastic and depends on the random initialization of the weights of the output linear classification layer, three training runs were carried out for each experiment. As a result, we give the mean with the standard deviation.

The total training time (without preliminary experiments) was 80 hours. One run of the training process for the model with the largest amount of training data (all seven training corpora) was 7 hours 20 minutes.

The class imbalance of the review corpora (see Table 2) was the reason that the macro-averaged F1-score was used as the main performance metric, which equally took into account the metrics for all classes, regardless of the number of texts. In addition, this metric was used in other papers exploring these corpora.

To simulate the expansion of the training dataset, two series of experiments have been carried out.

In **the first series**, the increase in the number of corpora was as follows:

- at the first stage, two training corpora of the SentiRuEval-2015 (*SRV15_car_tr* and *SRV15_rest_tr*) were used as a combined training corpus;
- at the second stage, they were joined by RuReviews and train part of the Russian Hotel Reviews (*RuReviews* and *Hotel_tr*);
- at the third stage, three training corpora ROMIP-2011 (*R11_book_tr*, *R11_mov_tr* and *R11_cam_tr*) were added and as a result, all seven training corpora were used;

In **the second series**, the extension of the training dataset was carried out as follows:

- at the first stage, three training corpora of the ROMIP-2011 (*R11_book_tr*, *R11_mov_tr* and *R11_cam_tr*) were used as a combined training corpus;
- at the second stage, two training corpora of the SentiRuEval-2015 (*SRV15_car_tr* and *SRV15_rest_tr*) were added to them;
- at the third stage, they were joined by RuReviews and training part of the Russian Hotel Reviews (*RuReviews* and *Hotel_tr*) and as a result, all seven training corpora were used.

The third stage of both series is the same experiment (all seven training corpora).

At each stage of each series, the RuBERT model was fine-tuned with the above mentioned hyperparameters. The fine-tuned model was tested on all nine test corpora.

As a baseline, we used the results of fine-tuning of the RuBERT for a situation where the training dataset was the related training corpus for the test corpus. For example, for test corpora of book reviews ROMIP-2011 (*R11_book_te*) and ROMIP-2012 (*R12_book_te*) as the training dataset we used the training corpus of book reviews ROMIP-2011 (*R11_book_tr*), and for the test corpus of car reviews SentiRuEval-2015 (*SRV15_car_te*) the training dataset was the training corpus of car reviews SentiRuEval-2015 (*SRV15_car_tr*).

### 4.2 Results

The results of the experiments are shown in Tables 3 and 4 and in Figure 2.

| Set of corpora | Number of corpora | R11_book_te | R11_mov_te | R11_cam_te | R12_book_te | R12_mov_te | R12_cam_te | SRV15_car_te | SRV15_rest_te | Hotel_te | Average |
|---|---|---|---|---|---|---|---|---|---|---|---|
| Baseline | 1 | .745 ±.013 | **.762 ±.015** | .885 ±.030 | .648 ±.031 | **.698 ±.023** | **.672 ±.024** | .786 ±.050 | .885 ±.021 | .912 ±.005 | .777 ±.092 |
| SRV15_car_tr, SRV15_rest_tr | 2 | .525 ±.016 | .577 ±.025 | .728 ±.024 | .602 ±.022 | .622 ±.009 | .516 ±.016 | .867 ±.020 | **.916 ±.023** | .773 ±.020 | .681 ±.017 |
| SRV15_car_tr, SRV15_rest_tr, RuReviews, Hotel_tr | 4 | .522 ±.018 | .608 ±.026 | .884 ±.025 | .613 ±.023 | .661 ±.010 | .642 ±.007 | **.887 ±.010** | .910 ±.012 | **.915 ±.005** | .738 ±.005 |
| All corpora | 7 | **.841 ±.020** | .756 ±.022 | **.915 ±.011** | **.724 ±.034** | .684 ±.024 | .665 ±.030 | .869 ±.018 | .890 ±.014 | .914 ±.003 | **.806 ±.011** |

Table 3: First series of experiments, macro-averaged $F_1$-score (Mean ± Std Dev)

| Set of corpora | Number of corpora | R11_book_te | R11_mov_te | R11_cam_te | R12_book_te | R12_mov_te | R12_cam_te | SRV15_car_te | SRV15_rest_te | Hotel_te | Average |
|---|---|---|---|---|---|---|---|---|---|---|---|
| Baseline | 1 | .745 ±.013 | **.762 ±.015** | .885 ±.030 | .648 ±.031 | **.698 ±.023** | **.672 ±.024** | .786 ±.000 | .885 ±.021 | .912 ±.005 | .777 ±.092 |
| R11_book_tr, R11_mov_tr, R11_cam_tr | 3 | .804 ±.004 | .755 ±.013 | .885 ±.013 | .608 ±.041 | .693 ±.015 | .653 ±.012 | .787 ±.041 | .759 ±.010 | .810 ±.016 | .751 ±.010 |
| R11_book_tr, R11_mov_tr, R11_cam_tr, SRV15_car, SRV15_rest | 5 | .808 ±.048 | .750 ±.011 | .895 ±.021 | .614 ±.044 | .676 ±.020 | .658 ±.018 | .840 ±.008 | **.890 ±.044** | .822 ±.009 | .773 ±.007 |
| All corpora | 7 | **.841 ±.020** | .756 ±.022 | **.915 ±.011** | **.724 ±.034** | .684 ±.024 | .665 ±.030 | **.869 ±.018** | **.890 ±.014** | **.914 ±.003** | **.806 ±.011** |

Table 4: Second series of experiments, macro-averaged $F_1$-score (Mean ± Std Dev)

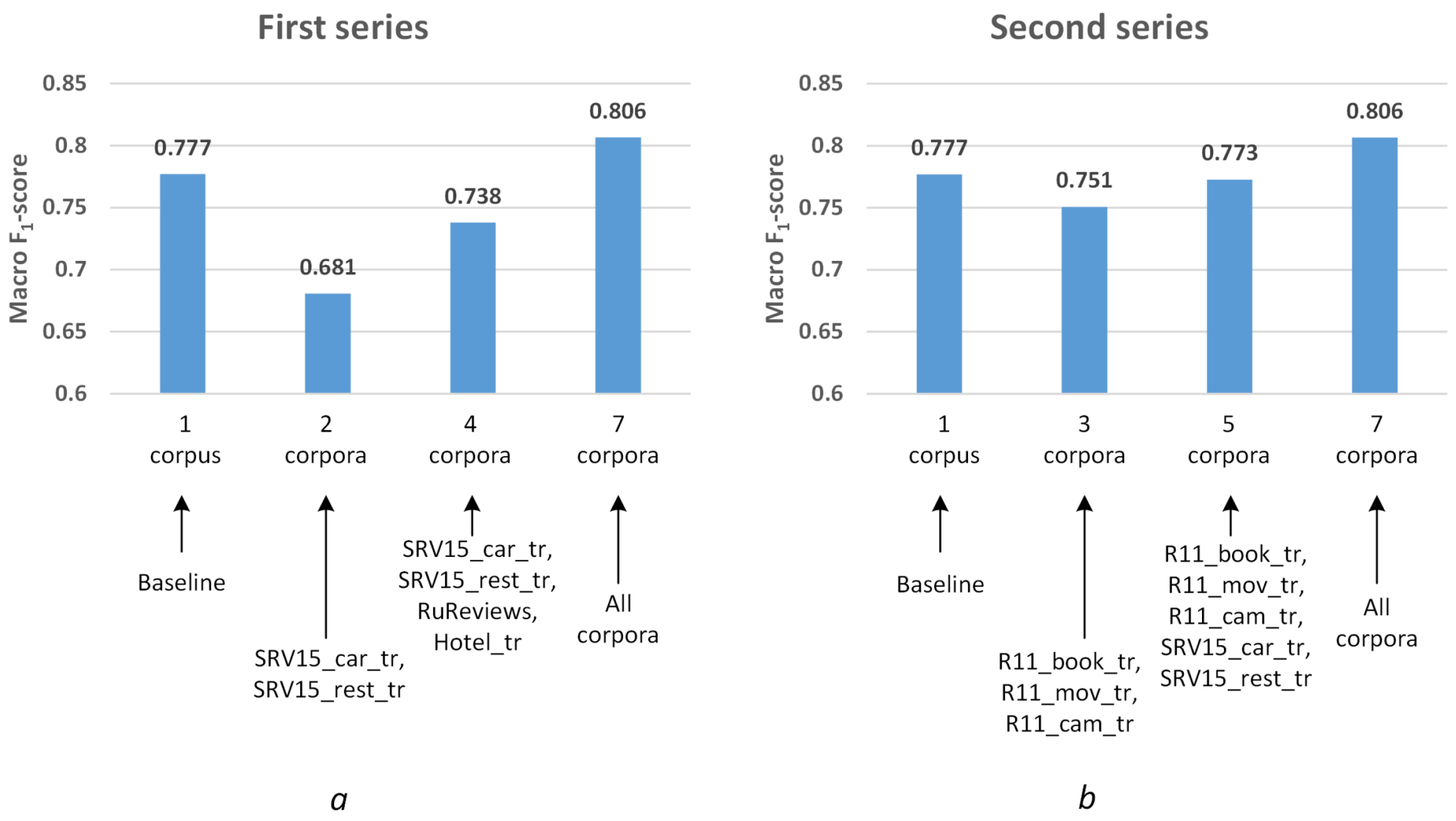

Figure 2: Macro-averaged $F_1$-scores for different numbers of training text corpora:
*a* – first experimental series, *b* – second experimental series

### 4.3 Discussion

For three corpora out of nine (*R11_book_te*, *R11_cam_te* and *R12_book_te*), as well as on average ($F_1$-score=0.806), the model trained on all seven training corpora shows the best result. It should be noted that this result has been obtained by one model, in contrast to the result in the second place (baseline: $F_1$-score=0.777), obtained by averaging the results of different models.

Models trained on small SentiRuEval-2015 corpora (second row of Table 3) show low results, except for the related test corpora (*SRV15_car_te* and *SRV15_rest_te*) – training data is clearly not enough for high-quality training. Adding training corpora RuReviews and Russian Hotel Reviews (third row of Table 3) significantly increases the average $F_1$-score (from 0.681 to 0.738), despite the fact that the domains of these two corpora do not correspond to the test corpora (with the exception of *Hotel_te*; but even excluding *Hotel_te* from consideration still gives an increase of the average $F_1$-score: from 0.681 to 0.716).

When training on all seven corpora (fourth row of Table 3) the performance scores for the SentiRuEval-2015 and Russian Hotel Reviews test corpora either do not decrease (for *SRV15_car_te* and *Hotel_te*), or decrease slightly (for *SRV15_rest_te*). Thus, the addition of ROMIP training corpora practically does not impair the learning process for these test corpora.

Models built on three ROMIP training corpora (second row of Table 4) for three of the six ROMIP test corpora (*R11_mov_te*, *R11_cam_te* and *R12_mov_te*) do not change the performance scores in comparison with training on the related corpora (baseline) (within 0.01), for two corpora reduce the performance scores (*R12_cam_te* – by 0.02 and *R12_book_te* – by 0.04) and for one corpus increase the performance score (*R11_book_te* – by 0.06).

The addition of SentiRuEval-2015 training corpora (third row of Table 4) significantly improves the performance for two SentiRuEval-2015 test corpora and practically does not change it for six ROMIP corpora.

Finally, the addition of RuReviews and Russian Hotel Reviews training corpora (fourth row of Table 4) significantly improves the performance for book, car and hotel corpora.

Figure 2 shows that in both series of experiments with an increase in the number of corpora (in the first series: 2 → 4 → 7; in the second series: 3 → 5 → 7) $F_1$-score on average increases. Thus, it can be concluded that expanding the training dataset has a positive effect on performance. In addition, the use of all available review corpora allows to obtain the best performance on average ($F_1$-score = 0.806). This circumstance allows us to look with cautious optimism at the possibility of building a universal neural network model for text sentiment analysis.

### 4.4 Comparison with previous results

Comparison of the obtained results with the results of other papers is possible only for the ROMIP-2011 and ROMIP-2012 test corpora. For SentiRuEval-2015, the performance scores are known for only three classes [11]; for Russian Hotel Reviews in [18] performance scores are given only for three aspects, but not for the review as a whole.

Table 5 shows the best results for the ROMIP-2011 test corpora from [4] and ROMIP-2012 from [5], as well as the results for these corpora obtained in our work: the results of the models trained on the related training corpora (our baseline – the first row in Tables 3 and 4) and the results of the model trained on all seven corpora (the last row in Tables 3 and 4). The best result for *R11_cam_te* in accordance with [4] was obtained using linear SVM; other two best models for the ROMIP-2011 test corpora were left unknown. The best result for *R12_book_te* in accordance with [5] was obtained by maximum entropy classifier, for *R12_mov_te* – rule-based classifier, and for *R12_cam_te* – linear SVM.

Neural network models for four corpora out of six have shown better results than the participants in the ROMIP competition. For the remaining two corpora, the results of neural network models differ from the previous results by less than 0.01. A significant advantage for the ROMIP-2011 book review corpus (0.841 vs 0.723) is due to the fact that the test corpus is highly imbalanced (244 positive reviews and 22 negative reviews – 91.7% and 8.3%), and the neural network model received a high macro precision (0.873 vs 0.698) due to accurate recognition of negative reviews ($precision_{pos}$=0.778).

| Model | R11_book_te | R11_mov_te | R11_cam_te | R12_book_te | R12_mov_te | R12_cam_te |
|---|---|---|---|---|---|---|
| The best models from [4] | 0.723 | **0.770** | **0.921** | – | – | – |
| The best models from [5] | – | – | – | 0.715 | 0.669 | 0.707 |
| The model trained on related training corpus (baseline) | 0.745 | 0.762 | 0.909 | 0.648 | **0.698** | **0.723** |
| The model trained on all the corpora | **0.841** | 0.756 | 0.915 | **0.724** | 0.684 | 0.665 |

Table 5: Results for test corpora of ROMIP-2011 and ROMIP-2012 (macro-averaged $F_1$-score)

## 5 Related work

Recently, several interesting papers have appeared in which the existing Russian-language corpora for the sentiment analysis have been investigated.

Zvonarev and Bilyi [26] used classifiers based on Logistic regression, XGBoost and Convolutional Neural Network for sentiment analysis of the RuTweetCorp. Baymurzina et al. [1] explored fastText and ELMo embeddings for sentiment analysis of the RuSentiment corpus.

At the end of 2018, the BERT neural network model [6] was presented based on the Transformer architecture [22], which showed State-of-the-Art results in several natural language processing tasks at once. After that, in several papers, the BERT model was applied for sentiment analysis in Russian.

Kuratov amd Arkhipov [10] built the RuBERT model – a Russian-language version of the BERT model based on the original multilingual version. RuBERT was used, inter alia, for sentiment analysis of the RuSentiment corpus.

Golubev and Loukachevitch [7] tested neural network models CNN, LSTM, BiLSTM and several variants of the BERT in the sentiment analysis task on the corpus of quotes ROMIP-2012[2], as well as on the SentiRuEval-2015 and SentiRuEval-2016 corpora.

Smetanin and Komarov [20] explored different versions of the BERT and Universal Sentence Encoder [3] for SentiRuEval-2015 (tweets), SentiRuEval-2016, RuTweetCorp, RuSentiment, Linis Crowd, Kaggle Russian News Dataset and RuReviews corpora.

In the above studies the state-of-the-art results for mentioned corpora were obtained on the basis of the BERT.

In our work, in contrast to those considered, we have investigated the effect of expanding the training dataset on the performance of sentiment analysis based on the BERT. In addition, in comparison with [20], our review includes the ROMIP-2011 and ROMIP-2012 corpora, and it was the first time that the performance scores of the BERT model have been obtained for them.

## 6 Conclusion

Currently, there are more than a dozen Russian-language text corpora, annotated by sentiment. These corpora differ significantly in sources, domains, sizes, quality of annotation and sentiment scales. Most of the corpora have a strong imbalance by classes, which reflects the distribution of texts in reality, but makes it difficult to train classifiers.

A variety of corpora can be used to build better models, which is confirmed by our experiments – the performance of the models increases on average with an increase in the number of training corpora. Also, information about the confidence level of the annotation quality can be used when choosing corpora for training and testing.

An important task is to study the possibility of constructing a universal sentiment analysis model that would find application in the fields where text analysis is required for many domains. In our work, it is shown that the performance is (obviously) strongly influenced by the presence of a corpus in a given domain. Less obvious was the fact that adding corpora in other domains, as a rule, either does not worsen the performance, or improves it.

Thus, the direction of further research is the possibility of building a universal model that is robust in relation to the domain.

[2] There is some uncertainty in the literature with the designation of the two ROMIP seminars held in 2011–2013: the first seminar is designated as ROMIP-2011 or ROMIP-2012, the second workshop – ROMIP-2012 or ROMIP-2013. This uncertainty stems from the fact that the first seminar was held in 2011, and the corresponding paper was published at the "Dialogue" conference in 2012. A similar situation took place with the second seminar.